\documentclass[
]{lib}

\usepackage{listings}
\usepackage{makecell}

\begin{document}

\copyrightyear{2025}
\copyrightclause{Copyright for this paper by its authors.
  Use permitted under Creative Commons License Attribution 4.0
  International (CC BY 4.0).}

\conference{}

\title{PABSA: Hybrid Framework for Persian Aspect-Based Sentiment Analysis}

\tnotemark[1]

\author[1]{Mehrzad Tareh}[%
orcid=0009-0008-4384-1463,
email=m.tareh@iasbs.ac.ir,
url=,
]

\author[1]{Aydin Mohandesi}[%
orcid=0009-0009-3136-9461,
email=mohandesi@iasbs.ac.ir,
url=,
]

\author[1]{Ebrahim Ansari}[%
orcid=0000-0002-3211-6679,
email=ansari@iasbs.ac.ir,
url=,
]
\cormark[1]

\address[1]{Department of CS and IT, Institute for Advanced Studies in Basic Sciences (IASBS), 
  Zanjan, Iran}

\cortext[1]{Corresponding author.}

\begin{abstract}
    Sentiment analysis is a key task in Natural Language Processing (NLP), enabling the extraction of meaningful insights from user opinions across various domains. However, performing sentiment analysis in Persian remains challenging due to the scarcity of labeled datasets, limited preprocessing tools, and the lack of high-quality embeddings and feature extraction methods. To address these limitations, we propose a hybrid approach that integrates machine learning (ML) and deep learning (DL) techniques for Persian aspect-based sentiment analysis (ABSA). In particular, we utilize polarity scores from multilingual BERT as additional features and incorporate them into a decision tree classifier, achieving an accuracy of 93.34\%—surpassing existing benchmarks on the Pars-ABSA dataset. Additionally, we introduce a Persian synonym and entity dictionary, a novel linguistic resource that supports text augmentation through synonym and named entity replacement. Our results demonstrate the effectiveness of hybrid modeling and feature augmentation in advancing sentiment analysis for low-resource languages such as Persian.
\end{abstract}

\begin{keywords}
    Persian aspect-based sentiment analysis \sep
    hybrid modeling \sep
    machine learning (ML) \sep
    deep learning (DL) \sep
    natural language processing (NLP) \sep
    polarity score \sep
    text augmentation \sep
    synonym and entity dictionary \sep
    low-resource languages
\end{keywords}
\maketitle

\section{Introduction}
Sentiment analysis is a fundamental task in NLP that aims to determine the sentiment or opinion expressed in a given text. It has widespread applications in areas such as customer feedback analysis, social media monitoring, product review, and market research. While sentiment analysis has been extensively studied for high-resource languages like English, Chinese, and French have been extensively researched \cite{tareh2025}, many low-resource languages, including Persian \cite{Ariai2024}, Amharic \cite{Tesfa2024}, African \cite{Chaudhary2023}, Arabic \cite{Bensoltane2021}, Hausa \cite{Musa2024} lack the necessary tools and resources for effective implementation.

ABSA is a more refined form of sentiment analysis that goes beyond general polarity detection by associating sentiments with specific aspects of a product, service, or entity. This is crucial in domains like e-commerce and customer reviews, where identifying sentiment toward particular attributes (e.g., price, quality, delivery, design, ...) provides more actionable insights. However, Pars-ABSA is still underexplored due to the limited availability of labeled datasets, embeddings, and feature extraction tools \cite{Shangipour2022}.

Despite the increasing use of Persian in digital communication, Persian sentiment analysis faces significant challenges. The primary issues include limited labeled datasets, as Persian lacks large-scale annotated datasets tailored for ABSA. Additionally, existing models trained on English or other languages perform poorly when applied to Persian due to linguistic differences. The morphological complexity of Persian, with its rich word formations and inflections, further complicates NLP processing. Another challenge is the lack of pre-trained embeddings and feature generators, as many widely used word embedding techniques are not optimized for Persian, limiting model effectiveness.

This study aims to address these challenges by developing and evaluating a hybrid ABSA model that integrates ML and DL techniques to improve sentiment classification in Persian. Various embeddings and feature extraction techniques tailored for Persian are explored, along with leveraging polarity scores from multilingual models to enhance sentiment classification. Furthermore, a Persian synonym and named-entity dictionary is introduced to support text augmentation and improve model performance.

This research makes several key contributions. First, it proposes a hybrid ABSA model that combines ML and DL approaches, surpassing previous state-of-the-art performance on the Pars-ABSA dataset. The model achieves an accuracy of 93.34\%, demonstrating the effectiveness of the methodology. Additionally, a Persian synonym and named-entity dictionary is developed, enabling synonym and named entity replacement for Persian NLP tasks. The study also benchmarks its approach against existing methods, including LSTM-based, AOA \cite{Huang2018}, Cabasc \cite{Liu2018}, RAM \cite{Chen2017}, and fine-tuned transformer models like Multilingual-BERT \cite{Jacob2018}, and ParsBERT \cite{Farahani2021}. By addressing the limitations of Persian sentiment analysis, this research contributes to advancing NLP for low-resource languages and provides valuable resources for future applications in Persian text processing.

\section{Related Work}
Sentiment analysis has been widely explored in high-resource languages such as English, where numerous datasets, tools, and models have been developed. Traditional sentiment analysis methods relied on lexicon-based \cite{Khoo2018} approaches and classical machine learning algorithms such as Support Vector Machines (SVM) and Naïve Bayes \cite{Alslaity2022}. More recently, deep learning models, including Convolutional Neural Networks (CNNs) and Long Short-Term Memory (LSTM) networks \cite{Rehman2019}, and transformer-based architectures \cite{Sayyida2022}, have demonstrated superior performance by capturing contextual information and complex linguistic patterns. For instance, a recent study reports a two-stage pipeline that first performs clustering and then applies a DeBERTa-based classifier for multi-label text categorization \cite{Chehreh2024}.

ABSA has gained traction as an essential subfield, allowing for fine-grained sentiment detection by associating opinions with specific aspects. The introduction of pre-trained transformer models, such as BERT \cite{Hoang2019} and its domain-specific variants, has significantly improved ABSA performance. In English and other widely spoken languages, datasets such as SemEval \cite{Pontiki2016} have facilitated advancements in this area. However, the lack of similar datasets and models for Persian limits the development of effective ABSA solutions for this language.

Persian sentiment analysis has primarily relied on traditional machine learning techniques and lexicon-based methods \cite{Basiri2018} due to the scarcity of labeled datasets and pre-trained models \cite{Vakili2024}. Early studies focused on rule-based and statistical approaches, with limited success in handling Persian's complex morphology and syntactic structures. The introduction of deep learning and transformer models, such as ParsBERT and Multilingual BERT (M-BERT), has improved sentiment classification accuracy \cite{Jafarian2021}. Despite these advancements, existing approaches often fail to achieve high performance due to the limited availability of high-quality annotated datasets and embeddings tailored for Persian \cite{Nazarizadeh2022}.

One of the main challenges in Persian NLP is the lack of comprehensive linguistic resources, such as sentiment lexicons, synonym dictionaries, and named entity datasets. Unlike English, where such resources are readily available, Persian remains a low-resource language, hindering the development of robust NLP models. Additionally, Persian exhibits complex morphological structures, making tokenization, stemming, and lemmatization more difficult. Another challenge is the limited generalizability of models trained on small datasets, as they often fail to capture the diverse linguistic variations present in Persian text.

Recent advances in ABSA have been driven by deep learning and transformer-based models. Pre-trained language models such as BERT, RoBERTa, and their multilingual counterparts have set new benchmarks in various NLP tasks \cite{Perikos2024}. Fine-tuning these models on domain-specific datasets has shown promising results in sentiment analysis. In Persian NLP, models like ParsBERT have been leveraged to improve sentiment classification. Additionally, feature engineering techniques, such as word embeddings including Word2Vec, Doc2Vec, FastText, GloVe, Tf-IDF, CountVectorizer \cite{Asudani2023}, and polarity score generation \cite{tareh2024}, have been explored to enhance model performance. However, there remains a gap in developing hybrid approaches that effectively integrate both ML and DL techniques to maximize accuracy and robustness in PABSA.

\section{Methodology}
ABSA aims to determine the sentiment polarity of specific aspects within a text. Unlike general sentiment analysis, which assigns an overall sentiment score, ABSA categorizes opinions based on predefined aspects such as product features or service quality. Each aspect is associated with a sentiment label, typically classified as positive, negative, or neutral. This fine-grained analysis is crucial in understanding nuanced user feedback, particularly in domains such as e-commerce and social media \cite{Nath2024}.

For this study, Persian text data was collected from various sources, including online reviews, social media comments, and Persian news websites. A significant portion of the dataset comes from Digikala, Iran’s largest e-commerce platform, where over 500,000 user reviews were gathered. The data underwent a manual annotation process, followed by validation by native Persian speakers to ensure accuracy. 

The data underwent multiple preprocessing steps, including tokenization, stopword removal, stemming, and normalization, to enhance text quality and improve model performance. Traditional NLP approaches, such as classical machine learning models (e.g., Naïve Bayes, SVM), were compared against deep learning-based methods, including CNNs, RNNs, and transformer models like BERT and ParsBERT. Transformer-based architectures demonstrated superior performance due to their ability to capture contextual dependencies and semantic nuances in Persian text.

Feature extraction played a crucial role in improving classification accuracy. Various word embedding techniques, including Word2Vec, FastText, and contextual embeddings from BERT, were evaluated for their effectiveness in aspect extraction. The polarity scores generated from DigiKala-BERT, Snapp-BERT, M-BERT were integrated as additional features to refine sentiment classification.

The models were trained and evaluated using standard metrics such as accuracy, F1-score, precision, and recall. The dataset was split into training, validation, and test sets to ensure robust evaluation and avoid overfitting. By leveraging a hybrid ML-DL approach and incorporating novel feature extraction techniques, this study achieved state-of-the-art performance in Persian ABSA, surpassing previous benchmarks.

\section{Results}
Our experimental setup utilized high-performance computing resources, including GPUs, to train and evaluate deep learning models. The models were implemented using TensorFlow and PyTorch, with extensive hyperparameter tuning to optimize factors such as batch size, learning rate, and optimizer selection.

To assess our approach, we compared it against existing Persian sentiment analysis models, including traditional ML methods and fine-tuned transformer-based models. The results demonstrated that our hybrid model significantly outperformed previous benchmarks, achieving an accuracy of 93.34\%, the highest recorded on the Pars-ABSA dataset.

Additionally, an ablation study was conducted to analyze the impact of different embeddings, model structures, and feature extraction techniques. The results revealed that combining multilingual BERT polarity scores with a decision tree classifier led to substantial performance improvements, validating the effectiveness of our approach.

\subsection{Experimental Setup and System Configuration}

The hardware configuration employed for both model training and evaluation is detailed in Table~\ref{tab:system_specifications}.  

\begin{table*}[h]
  \centering
  \caption{System Specifications}
  \label{tab:system_specifications}
  \begin{tabular}{ll}
    \toprule
    \textbf{Component} & \textbf{Specification} \\
    \midrule
    CPU & Intel(R) Xeon(R) E5-2620 v4 @ 2.10\,GHz \\
    CPU Cores & 8 cores per socket, 16 threads total \\
    RAM & 16\,GB (128\,MB block size) \\
    GPU & NVIDIA RTX 2080 Rev. A (8\,GB) \\
    Operating System & Linux-based environment \\
    \bottomrule
  \end{tabular}
\end{table*}

To maximize training efficiency and generalization performance, we conducted an extensive grid search to identify optimal hyperparameter values, as summarized in Table~\ref{tab:hyperparameters_values}. The selection process prioritized convergence stability, prevention of overfitting, and computational efficiency. Notably, the chosen learning rate, batch size, and warmup strategy contributed significantly to training stability and consistent performance across evaluation datasets.  

\begin{table*}[h]
  \centering
  \caption{Optimized Hyperparameter Settings}
  \label{tab:hyperparameters_values}
  \begin{tabular}{ll}
    \toprule
    \textbf{Hyperparameter} & \textbf{Value} \\
    \midrule
    Random Seed & 42 \\
    Batch Size & 16 \\
    Weight Decay & 0.01 \\
    Learning Rate & $2\times 10^{-5}$ \\
    Warmup Ratio & 0.06 \\
    Warmup Steps & 500 \\
    Maximum Sequence Length & 128 \\
    Training Epochs & 11 \\
    \bottomrule
  \end{tabular}
\end{table*}

\subsection{Performance Comparison}

We evaluated our proposed method against prior PABSA models developed since 2019. As summarized in Table~\ref{tab:sentiment_details}, the hybrid approach achieves substantial improvements in both accuracy and F1-score, outperforming all baselines. The most pronounced gains are observed over LCF-BERT (2019), reflecting advances in model architectures, preprocessing pipelines, and domain-specific fine-tuning. Notably, integrating contextualized embeddings with handcrafted linguistic features yields consistent performance gains over transformer-only models.

\begin{table*}[h]
  \centering
  \caption{Performance Comparison of Persian Sentiment Analysis Models}
  \label{tab:sentiment_details}
  \begin{tabular}{lccc}
    \toprule
    \textbf{Year} & \textbf{Approach} & \textbf{Accuracy (\%)} & \textbf{F1-Score (\%)} \\
    \midrule
    \textbf{2025} & \textbf{Hybrid Method} & \textbf{93.34} & \textbf{92.00} \\
    2022 & ParsBERT & 91.00 & 90.00 \\
    2019 & LCF-BERT & 87.40 & 86.30 \\
    \bottomrule
  \end{tabular}
\end{table*}

The results confirm that the M-BERT + Decision Tree configuration achieves state-of-the-art performance for Persian sentiment classification. The effectiveness stems from leveraging transformer-based embeddings to capture deep contextual semantics, combined with the structured decision boundaries of tree-based classifiers.

\subsection{Dataset Analysis}
Dataset characteristics are summarized in Table~\ref{tab:pars-absa-properties}, while Table~\ref{tab:pabsa_details} provides complementary statistical details such as sentiment class distribution and text length statistics. Finally, Table~\ref{tab:pars-absa-samples} presents representative annotated samples from the dataset. Following the protocol in prior studies, the dataset was randomly split into 80\% for training and 20\% for testing. All baseline and proposed models were trained and evaluated using this identical split to ensure a fair, directly comparable evaluation.

\begin{table*}[h]
  \centering
  \caption{Properties of Pars-ABSA Dataset}
  \label{tab:pars-absa-properties}
  \begin{tabular}{lc}
    \toprule
    \textbf{Property} & \textbf{Value} \\
    \midrule
    Number of sentiment targets & 10,002 \\
    Positive polarity targets & 5,114 \\
    Negative polarity targets & 3,061 \\
    Neutral polarity targets & 1,827 \\
    Total number of tokens & 693,825 \\
    Unique words & 18,270 \\
    Total number of comments & 5,602 \\
    Average words per comment & 123.85 \\
    \bottomrule
  \end{tabular}
\end{table*}
\begin{table*}[h]
\vspace{-0.5cm}
  \centering
  \caption{Statistical Details of the Pars-ABSA Dataset}
  \label{tab:pabsa_details}
  \begin{tabular}{lc}
    \toprule
    \textbf{Statistics} & \textbf{Information}\\ 
    \midrule
    Total Samples & 5,602 \\
    Total Aspects & 10,002 \\
    Positive & 5,114 (51.1\%) \\
    Neutral & 1,827 (18.3\%) \\
    Negative & 3,061 (30.6\%) \\
    Language & Persian \\
    Text Length & Avg: 293, Max: 5,238, Min: 9 \\
    \bottomrule
  \end{tabular}
\end{table*}
\begin{table}[hb]
\centering
\caption{Sample Reviews from the Pars-ABSA Dataset}
\includegraphics[width=\linewidth]{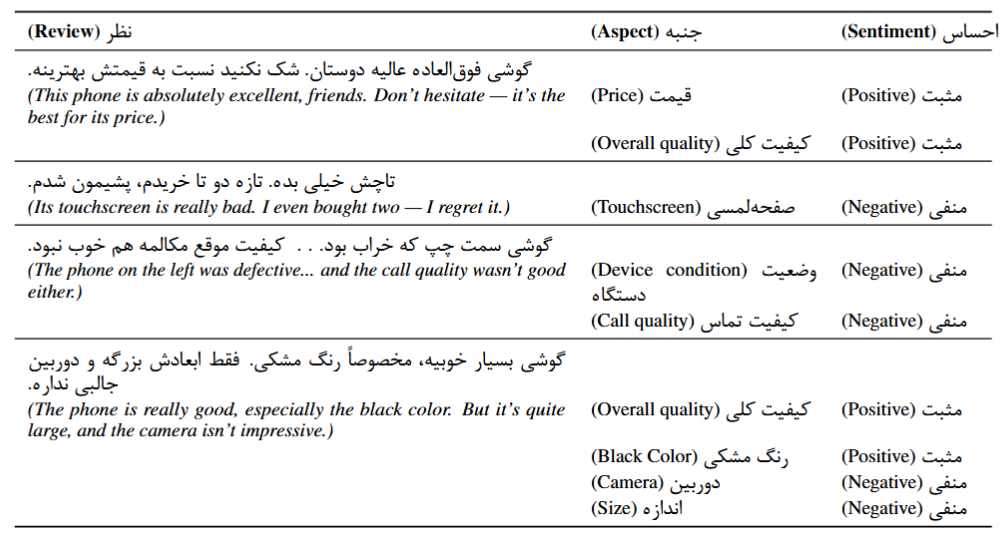}
\label{tab:pars-absa-samples}
\end{table}
\footnotesize\textit{Note:} The original reviews are in Persian (Farsi). They have been translated into English for clarity, while preserving sentiment labels and structure.

\newpage

\subsection{Persian Named Entity dataset}
The Persian Named Entity dataset consists of various categories of named entities in Persian. To enhance readability and global accessibility, we have translated the entity labels and examples into English while preserving the original structure and meaning.

Table \ref{tab:perner} provides an overview of different entity types along with sample instances from the dataset.

\begin{table}[h]
\centering
\caption{Persian Named Entities}
\includegraphics[width=\linewidth]{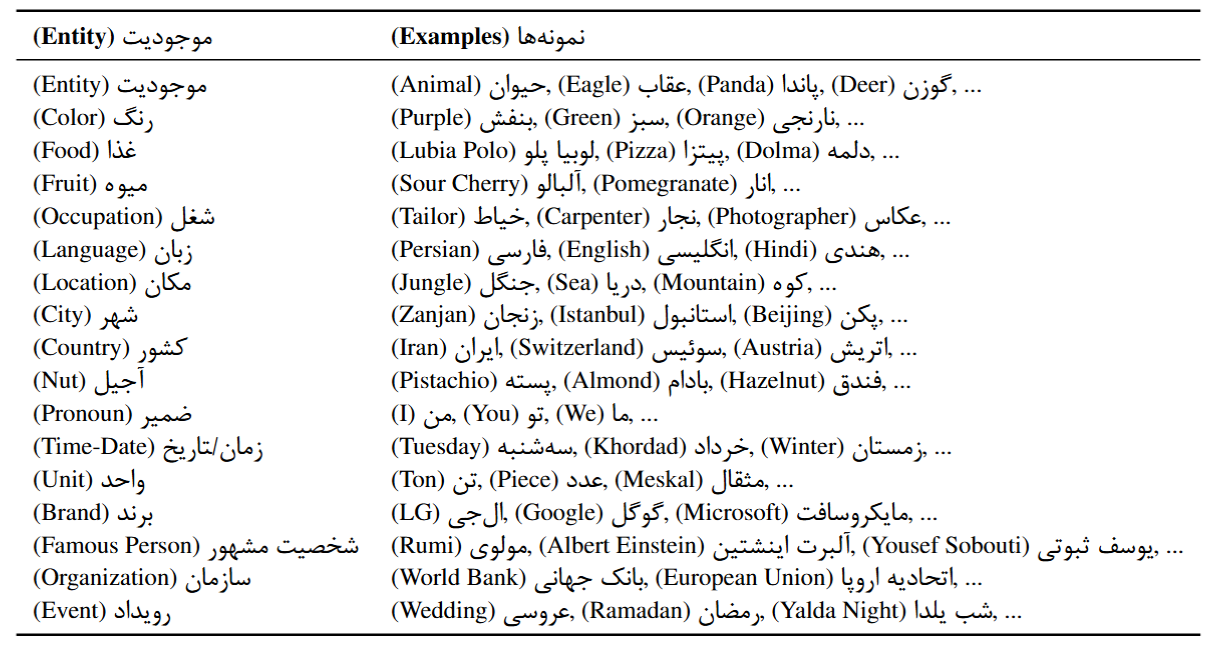}
\label{tab:perner}
\end{table}

\subsection{Persian Synonyms Vocabulary}
The Persian Synonyms Vocabulary dataset contains a collection of words along with their synonymous equivalents in Persian. To improve readability and accessibility, we have translated the words and their synonyms into English while maintaining their original meaning and structure. Table \ref{tab:persyns} presents examples of words with their corresponding synonyms.

\begin{table}[h]
\centering
\caption{Persian Synonyms}
\includegraphics[width=0.9\linewidth]{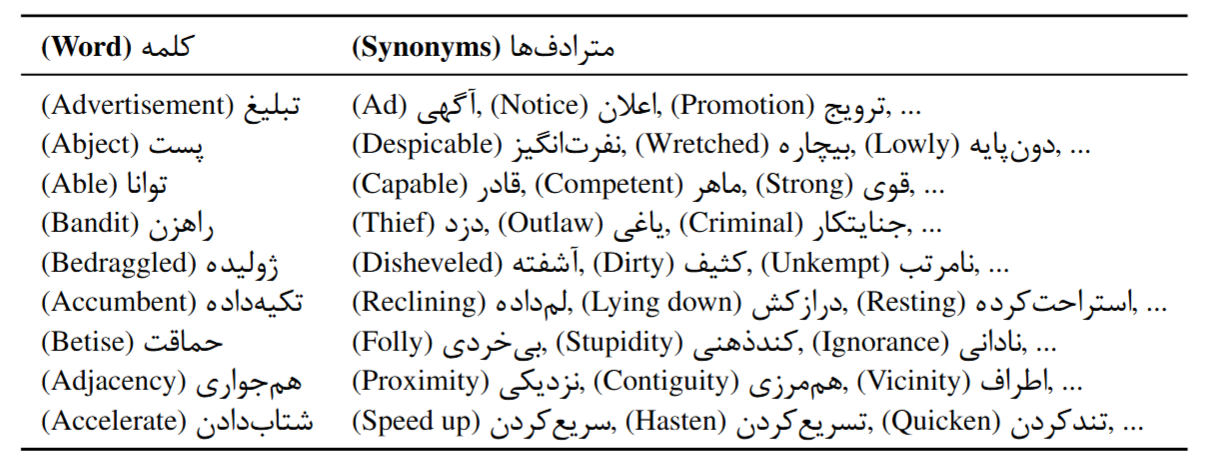}
\label{tab:persyns}
\end{table}

Since we now have access to Persian synonyms and named entities, we can apply synonym replacement and entity replacement techniques to our dataset in the context of ABSA. By strategically limiting the domain of the data that the model learns from, we enhance its ability to identify sentiment more accurately, handle linguistic variations, and improve generalization. These techniques not only aid in data augmentation but also help in reducing overfitting and making the model more adaptable to different expressions of sentiment in Persian text.

\section{Discussion}

The experimental results provide clear evidence of the advantages of hybrid architectures for PABSA. Specifically, the proposed \textit{M-BERT + Decision Tree} configuration consistently outperformed prior transformer-only methods, setting a new benchmark accuracy of 93.34\% on the Pars-ABSA dataset. This performance gain highlights the complementary strengths of transformer-based contextual embeddings and tree-based classifiers in capturing nuanced sentiment cues in low-resource languages.

Our findings also underscore the importance of domain-specific enhancements in Persian NLP. The incorporation of multilingual BERT--derived polarity scores as auxiliary features proved particularly impactful, offering richer semantic representations that improved classification robustness. Moreover, the introduction of Persian-specific linguistic resources---such as a synonym dictionary and named entity list---enabled effective data augmentation through synonym and entity replacement, reducing overfitting and improving generalization.

Despite these advances, several challenges persist. Persian’s complex morphology, scarcity of high-quality annotated corpora, and limited availability of domain-adapted embeddings remain significant barriers. Addressing these gaps through larger and more diverse datasets, advanced domain adaptation techniques, and richer feature engineering strategies could further elevate performance.

From an application perspective, the results have direct implications for e-commerce analytics, customer feedback mining, and social media monitoring in Persian-language contexts \cite{Saadati2024}. By enabling more accurate, context-aware sentiment classification, the proposed framework offers both academic and industrial value, laying the groundwork for more sophisticated Persian NLP solutions.

\section{Conclusion and Future Work}
This study introduced a hybrid approach for Persian aspect-based sentiment analysis, demonstrating superior performance over existing methods. Future research could focus on expanding labeled datasets, fine-tuning models for domain-specific applications, and integrating more advanced transformer architectures to further enhance performance. Additionally, addressing linguistic challenges such as sarcasm detection, ambiguity resolution, and irony recognition could significantly improve sentiment classification accuracy. These aspects are particularly crucial for analyzing social media comments, product reviews, and informal conversations, where sentiment is often implicit or context-dependent. The proposed model has potential applications in e-commerce, customer feedback analysis, and automated sentiment monitoring systems, contributing to advancements in Persian NLP and enabling more accurate and context-aware sentiment analysis.

\bibliography{references}

\end{document}